# Eye-Tracking Evolutionary Algorithm to minimize user fatigue in IEC applied to Interactive One-Max problem

Denis PALLEZ
LIRIS Lab
University of Lyon1
Lyon – France
denis.pallez
@unice.fr

Philippe COLLARD
I3S Lab
University of Nice
Nice – France
philippe.collard
@unice.fr

Thierry BACCINO
LPEQ Lab
University of Nice
Nice – France
thierry.baccino
@unice.fr

Laurent DUMERCY
LPEQ Lab
University of Nice
Nice – France
laurent.dumercy
@unice.fr

## ABSTRACT
In this paper, we describe a new algorithm that consists in combining an eye-tracker for minimizing the fatigue of a user during the evaluation process of Interactive Evolutionary Computation. The approach is then applied to the Interactive One-Max optimization problem.

## Categories and Subject Descriptors
D.3.3 **[Programming Languages]**: Language Contructs and Features – *abstract data types, polymorphism, control structures*.

I.2.10 **[Artificial Intelligence]**: Vision and Scene Understanding – *Perceptual reasoning*

J.4 **[Social And Behavioral Sciences]**: Psychology

## 1. INTRODUCTION
Interactive Evolutionary Computation (IEC) often suffers from user fatigue. In this paper, we present a new technique, totally independent of the domain used, to minimize this fatigue by combining an IEC and an input device. This device allows capturing where the user is looking on a monitor on which individuals are presented. This is possible by using eye-tracking systems such as Tobii™ which are totally non-intrusive for the user. Thus, we ensure there is no need for explicit user action (choosing and clicking the most promising individual, evaluating all the solutions etc.) during the evaluation process of the IEC; he just has to watch the screen and the presented individuals and to tell when he has finished evaluating/looking. The evolutionary algorithm then determines automatically which presented individuals are better by combining parameters obtained by a Tobii™ for each presented individual. We have applied to the Interactive One-Max problem [3]. Thus, by using totally implicit evaluation, we minimize the fatigue of the user in interactive computation, independently of the problem to be optimized. This approach may be used in any computer graphics application in which optimization or decision making is used.

In this paper, we first present related work in Interactive Evolutionary Computation, as well as an eye-tracking system and how it can be used with evolutionary algorithms. Next, we present the application we have developed to simulate this approach (Interactive One-max problem). We finish by presenting some results and future work.

## 2. IEC RELATED WORK
IEC is an optimization technique based on evolutionary computation (genetic algorithm, genetic programming, evolution strategy, or evolutionary programming) and used when it is hard or impossible to formalize efficiently the fitness function (the method that gives the performance of a solution to a given problem) and where the fitness function is therefore replaced by a human user. For instance, IEC is often used for optimization of subjective criteria such as aesthetics. A large survey of more than 250 papers can be obtained in [16], but the generally accepted first work on IEC is Dawkins [5], who studied the evolution of creatures called "biomorphs" by selecting them manually. Subsequently, much work was done in the area of computer graphics: for instance using IEC for optimizing lighting conditions for a given impression [1], applied to fashion design [9], or transforming drawing sketches into 3D models represented by superquadric functions and implicit surfaces, and evolving them by using divergence operators (bending, twisting, shearing, tapering) to modify the input drawing in order to converge to more satisfactory 3D pieces [12]. We can also mention work in combining human interactions with an artificial ant, applied to non-photorealistic rendering [15]. Another use of IEC involves a human patient using a PDA on which an IEC is launched to define best parameter values for cochlear implants [2]. First results show that patients using PDAs obtain a better parameterization than previously through lengthy interaction with a doctor. Following the same idea of using other human senses for human interaction, we can also mention the optimization of coffee blends [7].

As mentioned before, IEC is used when a fitness function is difficult and sometimes impossible to formalize. Human-Based Genetic Algorithms (HBGA) go further by allowing evolutionary computation where a good representation of individuals is hard or impossible to find [3], for instance they can be used in storytelling or in development of marketing slogans. To prove the usefulness of such techniques, the authors changed the classical One-Max optimization problem into an interactive one by interpreting the individuals (strings of bits – 0 or 1) as colors to be interactively presented and manipulated. We use the same approach to test our proposition.

Characteristics of IEC are *inconsistencies* of individuals fitness values given by the user, *slowness* of the evolutionary computation due to the interactivity, and *fatigue* of the user due to the obligation to evaluate manually all the individuals of each generation [14, 16]. For instance, most often the user is asked to give a mark to each individual or to select the most promising individuals according: it still requires active time consuming participation during the interaction. The number of individuals of a classical IEC is about 20 (the maximum that can be represented on the screen), and about the same for the number of generations.

However, some tricks are used to overcome those limits, e.g., trying to accelerate the convergence of IEC by showing the fitness landscape mapped in 2D or 3D, and by asking the user to determine where the IEC should search for a better optimum [6]. Other work tries to predict fitness values of new individuals based on previous subjective evaluation. This can be done either by constructing and approaching the subjective fitness function of the user by using genetic programming [4] or neural networks, or also with Support Vector Machine [10, 11]. In the latter case, inconsistent responses can also be detected thanks to graph based modeling.

Nonetheless, previous work is mostly algorithmic-oriented and not really user-oriented, which seems to be the future domain for IEC [13, 16]. In the next section, we will present material that can be combined with Interactive Evolutionary Computation in order to significantly reduce the active participation of the user during the evaluation process and to consequently reduce considerably the fatigue of the user and the slowness of IEC approaches.

## 3. EYE-TRACKING EVOLUTIONARY ALGORITHM (E-TEA)

### 3.1 What is an eye-tracking system?

An eye-tracking system consists of following the eye's motions while a user watches a screen on which something is presented. It pinpoints in real time the position where the eye is looking, with the help of one or two video cameras focusing on a reflected infrared ray sent to the user's cornea (cf. Figure 1). This device coupled with a computer regularly samples the space position of the eye and the pupil diameter. This latter parameter lets us know the cognitive intensity of the user: the more the user is concentrated on looking at something, the smaller the diameter [8]. Nowadays, eye-tracking systems are very useful because they can analyze in real time what a user is focused on without any effort and in a completely non-restrictive manner, in fact, the user does not know he is being observed by the machine. With such equipment, one can finally capture when, how much time, and with which cognitive intensity a screen area is looked at.

### 3.2 How to use an eye-tracker in IEC?

If we consider that either phenotype or genotype of individuals are graphically displayable on a screen, we can easily envisage using an eye-tracker during the evaluation process of IEC. Our proposal consists in using this hypothesis: the more an individual is examined, the better the fitness of this particular individual will be. So, a new evolutionary algorithm called Eye-Tracking Evolutionary Algorithm (E-TEA) is proposed:

1. generate initial population;
2. present the population to the user;
3. let the user watch the individuals

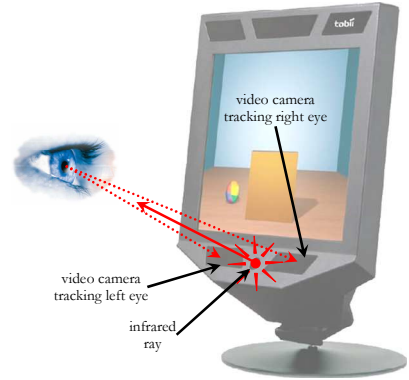

**Figure 1: How works an eye-tracker like Tobii™ ?**

4. compute how much time, how many times and with which cognitive intensity the presented individuals are looked at thanks to an eye-tracker;
5. combine previously obtained parameters and compute a fitness for each individual;
6. select the most promising individuals from the computed fitness
7. make crossover and mutation
8. return to step 2 until no further good individuals are found

Thus, the user just has to watch the screen and say when he has finished watching/evaluating. There is no need for the user to mark each individual, nor to choose the best or the most promising one. This will save considerable time and the user will be capable evaluating more solutions consequently there will be more evaluated generations. At a minimum, we estimate to double the number of generations. The principal difficulty is to determine how to combine different parameters obtained by the eye-tracker in order to define a computable fitness.

### 3.3 Estimated fitness formalization

As seen in previous sections, an eye-tracker like Tobii™ is able to provide at least 3 parameters for a screen region:

– let $d$ be the time the user has focused on a screen region;
– let $t$ be the number of transitions towards a particular screen region;
– let $p$ be the average of the pupil diameter when the user has focused on a screen region.

If we consider a screen region as an individual and if we suppose that the more an individual is observed, the better will be its fitness, we can define an estimated fitness of the region as:

(1) $$\hat{f}_u = \alpha d + \beta t + \gamma p$$

Unfortunately, $\alpha$, $\beta$, $\gamma$ values have to be defined empirically. In order to verify our hypothesis, we have conducted some experiments.

## 4. APPLICATION TO THE INTERACTIVE ONE-MAX OPTIMIZATION PROBLEM

Our optimization problem will be borrowed from [3] where the One-Max problem is considered as an interactive optimization problem in order to compare Interactive Genetic Algorithm (IGA) and Human-Based Genetic Algorithm (HBGA), and also in order to demonstrate the advantages of using HBGA. Recall that the classical One-Max optimization problem consists in maximizing the number of 1s in a string of bits (0 or 1). It is the simplest optimization problem and it is used here in order to parameterize our system. In the next paragraph, we will verify whether one-max optimization could be adapted to RGB colors. Then we present our interactive one-max problem.

### 4.1 One-max optimization vs. color optimization

In this section, we try to show that one-max optimization is rather equivalent to white color optimization in the RGB model even if it is not the best choice. Three distances for an objective fitness have been proposed [3]:

(2) $$M_1(R,G,B) = R + G + B$$

(3) $$M_2(R,G,B) = 255 \times \sqrt{3} - \sqrt{(255-R)^2 + (255-G)^2 + (255-B)^2}$$

(4) $$M_S(R,G,B) = \min(R,G,B)$$

We have studied the fitness-distance-correlation between each of the previous distances and the Hamming distance (number of 1s in the string). With 4000 samples, we found that $FDC(M_1) \approx -0.59$, $FDC(M_2) \approx -0.57$ and $FDC(M_S) \approx -0.48$. This means that $M_1$, representing the brightness, or $M_2$, representing the Euclidean distance between the considered and the white colors, are both correlated. Thus, one-max optimization can be adapted to interactive optimization by choosing the brighter color.

### 4.2 Implementation

As an eye-tracker is still very expensive, we have simulated such equipment with the help of a mouse. In fact, we ask the user to move the mouse to where he is looking. We know this is tedious, but it is the only way to simulate a Tobii™. Unfortunately, it is impossible to obtain values of the third parameter $p$. However, we think it is not unreasonable as a test. With this restriction, we have developed an application in Java 1.6 based on the ECJ library[1]. Rather than optimizing the simple one-max problem, we have decided to show individuals as colors [3]. Individuals are represented by a string of 24 bits, 8 bits each for red, green and blue. As we capture simulated eye motion, the screen presents only 8 zones (one individual per zone) and no individual in the center of the screen as shown in Figure 2. We avoid presenting solutions in the center because eyes are naturally attracted to the center. Also, if the user wants to compare two solutions that are diametrically opposite, eyes are obliged to cross the center. Consequently, the number of transitions for the center will increase considerably and will disrupt the estimated fitness of the solution which could be in the center.

When the user estimates he has finished watching solutions of a generation, we give him the possibility to click on his preferred color among the 8 presented. In that case, the estimated fitness is empirically cubed. The user also has the possibility to choose none of them. Thus, in Figure 2, we can see that during only the first 9 iterations colors are converging towards brighter colors.

---
[1] http://www.cs.gmu.edu/~eclab/projects/ecj/

Consequently, the estimated fitness we used for the $j^{th}$ individual depends whether the user has chosen it and is defined as:

(5) $$\hat{f}_u(j) = chosen ? \sqrt[3]{\frac{t_j}{2\sum_{i=1..8} t_i} + \frac{d_j}{2\sum_{i=1..8} d_i}} : \frac{t_j}{2\sum_{i=1..8} t_i} + \frac{d_j}{2\sum_{i=1..8} d_i}$$

Equation (5) is equivalent to equation (1) but we have normalized $\hat{f}_u$ in [0,1]. If solution $j$ is chosen, the first term is used, otherwise the second term.

### 4.3 Results

For the moment, it is difficult to give significantly quantitative results in so far as the application developed is only restricted to the use of a mouse and movements the user would give to it in order to simulate an eye-tracker. It is tedious work, but, we can say that it is easier to only move the mouse than to choose and click on the most promising individuals, or to evaluate them. In the future, it should be faster because interactions would be only with the eyes of the user. We estimate doubling, at a minimum the number of iterations in the Interactive Evolutionary Computation exploring a larger search space.

## 5. DISCUSSIONS

The Eye-Tracking Evolutionary Algorithm is a very simple but very innovative proposition that is at the intersection of two different domains: computer and cognitive sciences. This approach presents many advantages:

- First, it is the first time that an eye-tracker takes a very active part in a computer application. More traditionally, eye-tracking systems are used for analyzing human behavior when looking at an image, a text, a 3D model, a webpage, etc.
- Second, with such a combination we automate interactive evaluation of individuals with no constraints for the user. The only thing he has to do is to watch individuals and to say when he has finished. There is no explicit task imposed on the user, and thus no additional fatigue.
- Next, such material is completely non-intrusive, i.e., the user could forget that he is being observed. Interactive evaluation is as natural as possible.
- Finally, by analyzing the cognitive activity of the user, we can easily detect when the user shows signs of fatigue. For

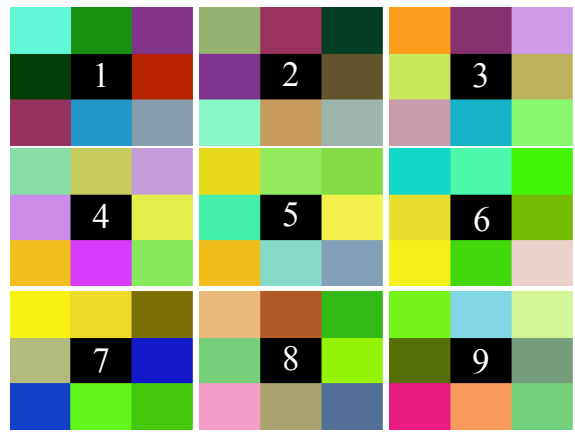

**Figure 2 : Screenshots of our Interactive One-max optimization problem (numbers represents the generation)**

instance, when the number of transitions between individuals is seriously decreasing or when the total time used to watch a generation is also decreasing, there is a chance that the user is bored. A pause can be made and the interactive evolutionary algorithm can be resumed later. However, the time used to watch individuals could be interpreted differently: the user is quickly converging toward a very good solution. More research has to be done to detect this fatigue.

Of course, each new system has its drawbacks, but they are few compared to the advantages:

– The eye-tracker can follow eyes if and only if it has been calibrated to the user. However, this takes only few seconds, and the user just has to focus on concentric moving circles.
– The other small constraint is that the user does not have total freedom of head movement. For instance, he can not look away and then resume evaluating. However, the freedom is large enough (30x16x20 cm) because of the use of two video cameras. If the signal is lost for one eye, the eye-tracker uses the other eye.

## 6. CONCLUSION & FUTURE WORK

In this article, we have presented a new algorithm that should considerably improve the speed of Interactive Evolutionary Computation. To do so, we have presented the Eye-Tracking Evolutionary Algorithm (E-TEA) that uses an eye-tracker in order to minimize user interaction for evaluating individuals. We have tested the approach by simulating an eye-tracker with a mouse during an interactive one-max optimization problem. The user had to move the mouse exactly to where he is interested by an individual. The only difference with a real eye-tracker is the loss of crucial information about cognitive intensity represented by the pupil diameter. Nonetheless, we are convinced that time taken during the evaluation process can be significantly reduced.

In the future, we will first create an application interfacing the interactive one-max problem and a real eye-tracker in order to correctly parameterize our interactive evolutionary algorithm. Next, we want to test it on a real world application.

## 7. ACKNOWLEDGMENTS

We would like to thank the Institute of Technology at Nice University (http://www.iut-nice.fr) which let this research work be possible and also to thank Pr. Peter Sander for his precious help.